\documentclass{article}
\usepackage{graphicx}
\usepackage{graphicx}
\usepackage{booktabs}
\usepackage{dsfont}
\usepackage{multicol}
\usepackage{multirow}
\usepackage{adjustbox}
\usepackage{makecell}
\usepackage{color, colortbl}
\usepackage[bookmarks=false]{hyperref}
\usepackage{dirtytalk}
\usepackage{tablefootnote}
\hypersetup{
    colorlinks=true,
    linkcolor=blue,
    filecolor=magenta,      
    urlcolor=cyan,
    pdftitle={Overleaf Example},
    pdfpagemode=FullScreen,
    }

\usepackage{mathtools}
\usepackage[square,sort,comma,numbers]{natbib}
\newcommand\Myperm[2][^n]{\prescript{#1\mkern-2.5mu}{}P_{#2}}

\usepackage{wrapfig}
\usepackage{tikz}
\usepackage{comment}
\usepackage{amsmath,amssymb} 
\usepackage{color}
\usepackage{pifont}
\newcommand{\cmark}{\ding{51}}%
\newcommand{\xmark}{\ding{55}}%
\newcommand\methodname{Relationformer}
\definecolor{LightCyan}{rgb}{0.88,1,1}

\usepackage[preprint]{main}

\usepackage[utf8]{inputenc} 
\usepackage[T1]{fontenc}    
\usepackage{hyperref}       
\usepackage{url}            
\usepackage{booktabs}       
\usepackage{amsfonts}       
\usepackage{nicefrac}       
\usepackage{microtype}      

\title{\methodname: A Unified Framework for \textit{Image-to-Graph} Generation}

\author{Suprosanna Shit\thanks{equal contribution}  $~^{1,3}$, Rajat Koner\footnotemark[1] $~^{2}$, Bastian Wittmann$^{1}$, Johannes Paetzold$^{1}$,\\ \textbf{Ivan Ezhov$^{1}$, Hongwei Li$^{1,3}$, Jiazhen Pan$^{1}$, Sahand Sharifzadeh$^{2}$,} \\ \textbf{Georgios Kaissis$^{1}$, Volker Tresp$^{2}$, Bjoern Menze$^{1,3}$}\\
$^{1}$Technical University of Munich, $^{2}$Ludwig Maximilian University of Munich,\\
$^{3}$University of Zurich\\
\texttt{suprosanna.shit@tum.de, koner@dbs.ifi.lmu.de}\\
}

\begin{document}

\maketitle

\begin{abstract}
A comprehensive representation of an image requires understanding objects and their mutual relationship, especially in \textit{image-to-graph} generation, e.g., road network extraction, blood-vessel network extraction, or scene graph generation. Traditionally, \textit{image-to-graph} generation is addressed with a two-stage approach consisting of object detection followed by a separate relation prediction, which prevents simultaneous object-relation interaction. This work proposes a unified one-stage transformer-based framework, namely \methodname~that jointly predicts objects and their relations. We leverage direct set-based object prediction and incorporate the interaction among the objects to learn an object-relation representation jointly.
In addition to existing [\texttt{obj}]-tokens, we propose a novel learnable token, namely [\texttt{rln}]-token. Together with [\texttt{obj}]-tokens, [\texttt{rln}]-token exploits local and global semantic reasoning in an image through a series of mutual associations. In combination with the pair-wise [\texttt{obj}]-token, the [\texttt{rln}]-token contributes to a computationally efficient relation prediction. We achieve state-of-the-art performance on multiple, diverse and multi-domain datasets that demonstrate our approach's effectiveness and generalizability. \footnote{code is available at \url{https://github.com/suprosanna/relationformer}}
\end{abstract}
\section{Introduction}
\label{sec:intro}

An image contains multiple layers of abstractions, from low-level features to intermediate-level objects to high-level complex semantic relations. To gain a complete visual understanding, it is essential to investigate different abstraction layers jointly. An example of such multi-abstraction problem is \textit{image-to-graph} generation, such as road-network extraction \cite{he2020sat2graph}, blood vessel-graph extraction~\cite{paetzold2021whole}, and scene-graph generation \cite{xu2017scenegraph}. In all of these tasks, one needs to explore not only the objects or the \textit{nodes}, but also their mutual dependencies or relations as \textit{edges}.

\begin{figure}[t]
    \centering
    \includegraphics[width=1.0\textwidth]{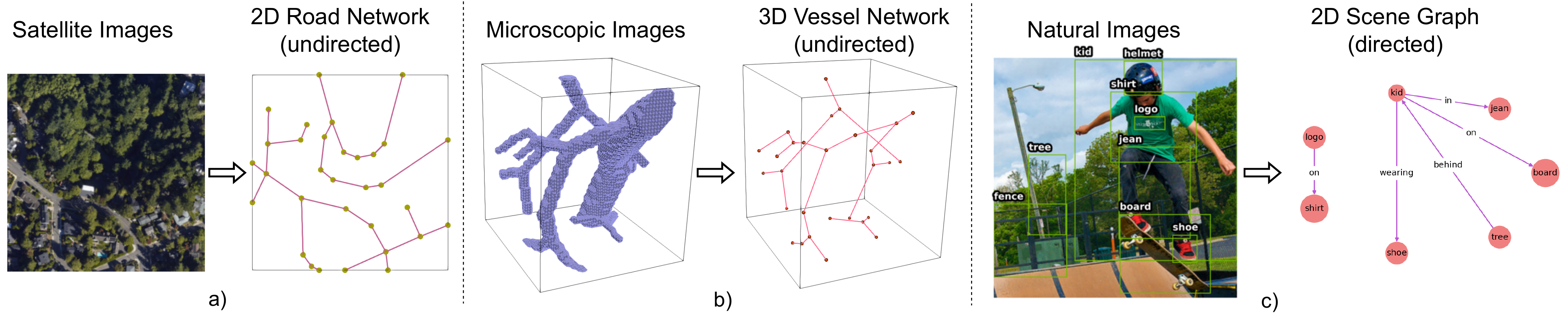}
    \vspace{-2em}
    \caption{Examples of relation prediction tasks. Note that the 2D road network extraction and 3D vessel graph extraction tasks have undirected relations while the scene graph generation task has directed relations.}
    \label{fig:example_prob}
    \vspace{-1.5em}
\end{figure}

In \textit{spatio-structural} tasks, such as road network extraction (Fig. \ref{fig:example_prob}a), nodes represent road-junctions or significant turns, while edges correspond to structural connections, i.e., the road itself. The resulting spatio-structural graph construction is crucial for navigation tasks, especially with regard to autonomous vehicles. Similarly, in 3D blood vessel-graph extraction (Fig. \ref{fig:example_prob}b), nodes represent branching points or substantial curves, and edges correspond to structural connections, i.e., arteries, veins, and capillaries. Biological studies relying on a vascular graph representation, such as detecting collaterals \cite{todorov2020machine}, assessing structural robustness \cite{ji2021brain}, emphasize the importance of efficient extraction thereof. In case of \textit{spatio-semantic} graph generation, e.g. scene graph generation from natural images (Fig. \ref{fig:example_prob}c), the objects denote nodes and the semantic-relation denotes the edges \cite{johnson2015image}. This graphical representation of natural images is compact, interpretable, and facilitates various downstream tasks like visual question answering \cite{hildebrandt2020scene,koner2021graphhopper}. Notably, different image-to-graph tasks have been addressed separately in previous literature (see Sec. \ref{sec:rel_lit}), and to the best of our knowledge, no unified approach has been reported so far.

Traditionally, image-to-graph generation has been studied as a complex multistage pipeline, which consist of an object detector \cite{ren2015faster}, followed by a separate relation predictor \cite{koner2020relation,lin2020gps}. Similarly, for spatio-structural graph generation, the usual first stage is segmentation, followed by a morphological operation on binary data. While a two-stage \textit{object-relation} graph generation approach is modular, it is usually trained sequentially, which increases model complexity and inference time and lacks simultaneous exploration of shared object-relation representations. Additionally, mistakes in the first stage may propagate into the later stages. It should also be noted that the two-stage approach depends on multiple hand-designed features, spatial \cite{zellers2018neural}, or multi-modal input \cite{chen2019knowledge}.

We argue that a single-stage image-to-graph model with joint object and relation exploration is efficient, faster, and easily extendable to multiple downstream tasks compared to a traditional multi-stage approach. Crucially, it reduces the number of components and simplifies the training and inference pipeline (Fig. \ref{fig:overview}). Furthermore, intuitively, a simultaneous exploration of objects and relations could utilize the surrounding context and their co-occurrence. For example, Fig. \ref{fig:example_prob}c depicting the \say{kid} \say{on} a \say{board} introduces a spatial and semantic inclination that it could be an outdoor scene where the presence of a \say{tree} or a \say{helmet}, the kid might wear, is highly likely. The same notion is analogous in a spatio-structural vessel graph. Detection of a \say{bifurcation point} and an \say{artery} would indicate the presence of another \say{artery} nearby. The mutual co-occurrence captured in joint object-relation representation overcomes individual object boundaries and leads to a more informed big picture.

Recently, there has been a surge of one-stage models in object detection thanks to the DETR approach described in  \cite{carion2020end}. These one-stage models are popular due to the simplicity and the elimination of reliance on hand-crafted designs or features. DETR exploits a encoder-decoder transformer architecture and learns object queries or [\texttt{obj}]-token for object representation. 

To this end, we propose \textbf{\methodname}, a \textit{unified one-stage} framework for end-to-end image-to-graph generation. We leverage set-based object detection of DETR and introduce a novel learnable token named [\texttt{rln}]-token in tandem with [\texttt{obj}]-tokens. The [\texttt{rln}]-token captures the inter-dependency and co-occurrence of low-level objects and high-level spatio-semantic relations. \methodname~directly predicts objects from the learned [\texttt{obj}]-tokens and classifies their pairwise relation from combinations of [\texttt{obj-rln-obj}]-tokens. In addition to capturing pairwise object-interactions, the [\texttt{rln}]-token, in conjunction with relation information, allows all relevant [\texttt{obj}]-tokens to be aware of the global semantic structure. These enriched [\texttt{obj}]-tokens in combination with the relation token, in turn, contributes to the relation prediction. The mutually shared representation of joint tokens serves as an excellent basis for an image-to-graph generation. Moreover, our approach significantly simplifies the underlying complex image-to-graph pipeline by only using image features extracted by its backbone.

We evaluate \methodname~across numerous publicly available datasets, namely Toulouse, 20 US Cities, DeepVesselNet, and Visual Genome, comprising 2D and 3D, directed- and undirected image-to-graph generation tasks. We achieve a new state-of-the-art for one-stage methods on Visual Genome, which is better or comparable to the two-stage approaches. We achieve state-of-the-art results on road-network extraction on the Toulouse and 20 US Cities dataset. To the best of our knowledge, this is the first image-to-graph approach working directly in \emph{3D}, which we use to extract graphs formed by blood vessels.

\section{Related Work}
\label{sec:rel_lit}

\begin{figure}[t!]
    \centering
    \includegraphics[width=1.0\textwidth]{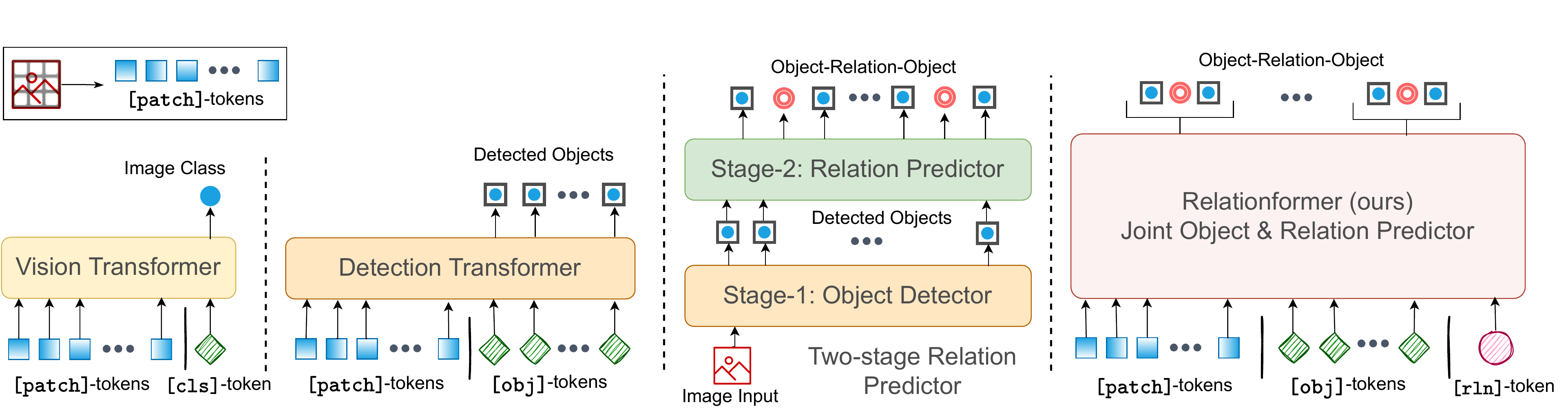}
    \scriptsize
    \caption{This illustrates a general architectural evolution of transformers in computer vision and how \methodname~advances the concept of a task-specific learnable token one step further. The proposed \methodname~is also shown in comparison to the conventional two-stage relation predictor. The amalgamation of two separate stages not only simplifies the architectural pipeline but also co-reinforces both of the tasks.}
    \label{fig:overview}
\end{figure}

\paragraph{\textbf{Transformer in Vision:}}  In recent times, transformer-based architectures have emerged as the de-facto gold standard model for various multi-domain and multi-modal tasks such as image classification \cite{dosovitskiy2020image}, object detection \cite{carion2020end}, and out-of-distribution detection \cite{koner2021oodformer}. DETR \cite{carion2020end} proposed an end-to-end transformer-based object detection approach with learnable object queries ([\texttt{obj}]-tokens) and direct set-based prediction. DETR eliminates burdensome object detection pipelines (e.g., anchor boxes, NMS) of traditional approaches \cite{ren2015faster} and directly predicts objects. Building on DETR, a series of object detection approaches improved DETR's slow convergence \cite{zhu2020deformable}, adapted a pure sequence-to-sequence approach \cite{fang2021you}, and improved detector efficiency \cite{song2021vidt}. In parallel, the development of the vision transformer \cite{dosovitskiy2020image} for image classification offered a powerful alternative. Several refined idea \cite{touvron2021training,liu2021swin} have advanced this breakthrough and transformer in general emerges as a cutting-edge research topic with focus on novel design principle and innovative application. Fig. \ref{fig:overview}, shows a pictorial overview of transformer-based image classifier, object detector, and relation predictor including our proposed method, which we referred to as \methodname.

\paragraph{\textbf{Spatio-structural Graph Generation:}}
In a spatio-structural graph, the most important physical objects are edges, i.e., roads for a road network or arteries and veins in vessel graphs. Conventionally, spatio-structural graph extraction has only been discussed in 2D with little-to-no attention on the 3D counterpart. For 2D road network extraction, the predominant approach is to segment \cite{mattyus2017deeproadmapper,batra2019improved} followed by morphological thinning to extract the spatial graph. Only few approaches combine graph level information processing, iterative node generation \cite{bastani2018roadtracer}, sequential generative modelling \cite{chu2019neural}, and graph-tensor-encoding \cite{he2020sat2graph}. 
Belli et al. \cite{belli2019image} for the first time, adopted attention mechanisms in an auto-regressive model to generate graphs conditioned on binary segmentation. Importantly, to this date, none of these 2D approaches has been shown to scale to 3D. 

For 3D vessel-network extraction, segmentation of whole-brain microscopy images \cite{todorov2020machine,miettinen2021micrometer} has been combined with rule-based graph extraction algorithms \cite{shit2021cldice}. Recently, a large-scale study \cite{paetzold2021whole} used the \textit{Voreen} \cite{meyer2009voreen} software to extract whole-brain vascular graph from binary segmentation, which required complicated heuristics and huge computational resources. Despite recent works on 3D scene graphs \cite{armeni20193d} and temporal scene graphs \cite{ji2020action}, to this day, there exists no learning-based solution for 3D spatio-structural graph extraction.

Considering two spatio-structural image-to-graph examples of vessel-graph and road-network, one can understand the spatial relation detection task as a link prediction task. In link-prediction, graph neural networks, such as GraphSAGE \cite{hamilton2017inductive}, SEAL \cite{zhang2018link} are trained to predict missing links among nodes using node features. These approaches predict links on a given set of nodes. Therefore, link prediction can only optimize correct graph topology. In comparison, we are interested in joint node-edge prediction, emphasizing correct topology and correct spatial location simultaneously, making the task even more challenging.

\paragraph{\textbf{Spatio-semantic Graph Generation:}} Scene graph generation (SGG) \cite{lu2016visual,xu2017scenegraph} from 2D natural images has long been studied to explore objects and their inter-dependencies in an interpretable way. Context refinement across objects \cite{xu2017scenegraph,zellers2018neural}, extra modality of features \cite{lu2016visual,sharifzadeh2021improving} or prior knowledge \cite{sharifzadeh2021classification} has been used to model inter-dependencies of objects for relation prediction. RTN \cite{koner2020relation,koner2021scenes} was one of the first transformer approaches to explore context modeling and interactions between objects and edges for SGG. Li et al. \cite{li2021sgtr} uses DETR like architecture to separately predict entity and predicate proposal followed by a graph assembly module. Later, several works \cite{dhingra2021bgt,lu2021context} explored transformers, improving relation predictions. On the downside, such two-stage approaches increase model size, lead to high inference times, and rely on extra features such as glove vector \cite{pennington2014glove} embedding or knowledge graph embedding \cite{sharifzadeh2022improving}, limiting their applicability. Recently, Liu et al. \cite{liu2021fully} proposed a fully convolutional one-stage SGG method. It combined a feature pyramid network \cite{lin2017feature} and a relation affinity field \cite{zhou2019objects,newell2017pixels} for modeling a joint \textit{object-relation} graph. However, their convolution-based architecture limits the context exploration across objects and relations. Contemporary to us \cite{cong2022reltr} used transformers for the task of SGG. However, their complex pipeline for a separate subject and object further increases computational complexity. Crucially, there has been a significant performance gap between one-stage and two-stage approaches. This paper bridges this gap with simultaneous contextual exploration across objects and relations.

\section{Methodology}
\label{sec:method}

In this section, we formally define the generalized \textit{image-to-graph} generation problem. Each of the presented relation prediction tasks in Figure \ref{fig:example_prob} is a special instance of this generalized image-to-graph problem. Consider an image space $I \in \mathbb{R}^{D \times \text{\#ch}}$, where $D=\prod_{i=1}^{d}\text{dim}[i]$ for a $d$ dimensional image and \text{\#ch} denotes the number of channels. Now, an image-to-graph generator $\mathcal{F}$ predicts $\mathcal{F}(I)=\mathcal{G}$ for a given image $I$, where $\mathcal{G}=(\mathcal{V}, \mathcal{E})$ represents a graph with vertices (or objects) $\mathcal{V}$ and edges (or relations) $\mathcal{E}$. Specifically, the $i^{\text{th}}$ vertex $v^i\in \mathcal{V}$ has a node or object location specified by a bounding box $\boldsymbol{v}^i_{\text{box}}\in \mathbb{R}^{2\times d}$ and a node or object label $v^i_{\text{cls}} \in \mathbb{Z}^C$. Similarly, each edge $e^{ij}\in \mathcal{E}$ has an edge or relation label $e^{ij}_{\text{rln}} \in \mathbb{Z}^L$, where we have $C$ number of object classes and $L$ types of relation classes. Note that $\mathcal{G}$ can be both directed and undirected. The algorithmic complexity of predicting graph $\mathcal{G}$ depends on its size, $|\mathcal{G}|=|\mathcal{V}|+|\mathcal{E}|$ which is of order complexity $\mathcal{O}(N^2)$ for $|\mathcal{V}|=N$. It should be noted that object detection as a special case of the generalized image-to-graph generation problem, where $\mathcal{E}=\phi$. In the following, we briefly revisit a set-based object detector before expanding on our rationale and proposed architecture.

\subsection{Preliminaries of Set-based Object Detector}
Carion et al. \cite{carion2020end} proposed DETR, which shows the potential of set-based object detection, building upon an encoder-decoder transformer architecture \cite{vaswani2017attention}. Given an input image $I$, a convolutional backbone \cite{he2016deep} is employed to extract high level and down scaled features. Next, the spatial dimensions of extracted features are reshaped into a vector to make them sequential. Afterwards, these sequential features are coupled with a sinusoidal positional encoding \cite{bello2019attention} to mark an unique position identifier. A stacked encoder layer, consisting of a multi-head self-attention and a feed-forward network, processes the sequential features. The decoder takes $N$ number of learnable object queries ([\texttt{obj}]-tokens) in the input sequence and combines them with the output of the encoder via cross-attention, where $N$ is larger than the maximum number of objects.

DETR utilizes the direct Hungarian set-based assignment for one-to-one matching between the ground truth and the predictions from $N$ [\texttt{obj}]-tokens. The bipartite matching assigns a unique predicted object from the $N$ predictions to each ground truth object. Only matched predictions are considered valid. The rest of the predictions are labeled as $\varnothing$ or `background'. Subsequently, it computes the box regression loss solely for valid predictions. For the classification loss, all predictions, including `background' objects, are considered.

In our work, we adopt a modified attention mechanism, namely deformable attention from deformable-DETR (def-DETR) \cite{zhu2020deformable} for its faster convergence and computational efficiency. In DETR, complete global attention allows each token to attend to all other tokens and hence capture the entire context in one image. However, information about the presence of an object is highly localized to a spatial position. Following the concept of deformable convolutions \cite{dai2017deformable}, deformable attention enables the queries to attend to a small set of spatial features determined from learned offsets of the reference points. This improves convergence and reduces the computational complexity of the attention operation. 

Let us consider an image feature map $\boldsymbol{f}_I$, the $q^{th}$ [\texttt{obj}]-token with associated features $\boldsymbol{f}_q$, and the reference point $\boldsymbol{x}_q$. First, for the $m^{th}$ attention head, we need to compute the $k^{th}$ sampling offset $\Delta\boldsymbol{x}_{mqk}$ based on the query features $\boldsymbol{f}_q$. Subsequently, the sampled image features $\boldsymbol{f}_I(\boldsymbol{x}_q + \Delta \boldsymbol{x}_{mqk})$ go through a single layer $\boldsymbol{W}_{m}^{'}$ followed by a multiplication with the attention weight $\boldsymbol{A}_{mqk}$, which is also obtained from the query features $\boldsymbol{f}_q$. Finally, another single layer $\boldsymbol{W}_m$ merges all the heads. Formally, the deformable attention operation (DefAttn) for $M$ heads and $K$ sampling points is defined as:
\begin{eqnarray}
    \text{DefAttn}(\boldsymbol{f}_q,\boldsymbol{x}_q,\boldsymbol{f}_I) = \sum_{m=1}^{M}\boldsymbol{W}_m\left[\sum_{k=1}^{K}\boldsymbol{A}_{mqk}\cdot\boldsymbol{W}_{m}^{'}\boldsymbol{f}_I(\boldsymbol{x}_q + \Delta \boldsymbol{x}_{mqk})\right]
\end{eqnarray}

\begin{figure*}[t]
    \centering
    \includegraphics[width=1.0\textwidth]{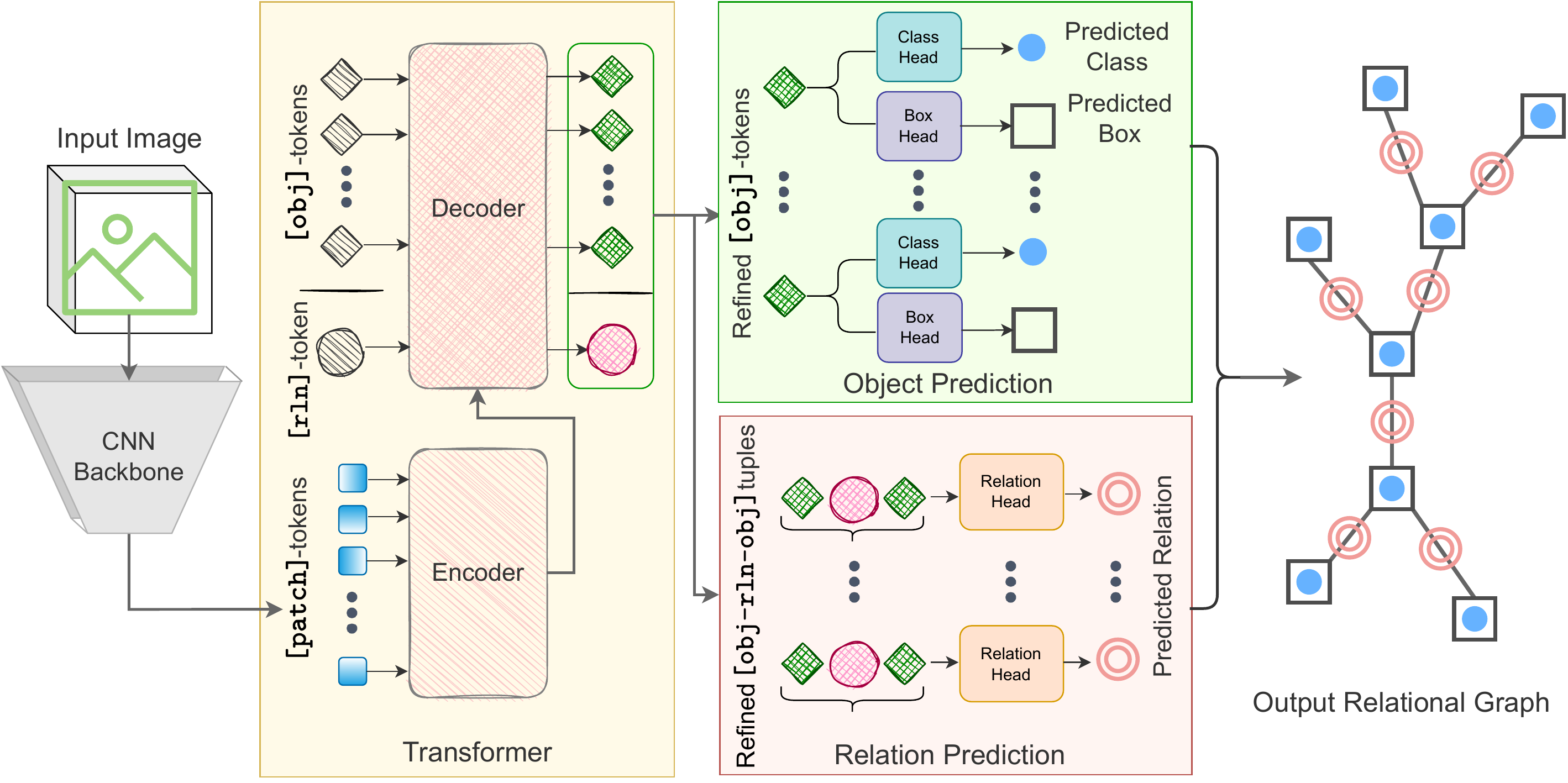}
    \vspace{-1em}
    \caption{Specifics of the \methodname~architecture. The image is first processed by a feature extractor, which generates [\texttt{patch}]-tokens for the input of the transformer encoder. Next, transformer decoder takes learnable [\texttt{obj}]-tokens and a [\texttt{rln}]-token along with output from encoder. Decoder processes them through a series of self- and cross-attention operations. The object head processes the final [\texttt{obj}]-tokens from the decoder to produce the bounding box and object classes. The relation head takes a tuple of the final [\texttt{obj-rln-obj}]-token combination and classifies their relation. Combining the output of the object and relation head yields the final graph.}
    \label{fig:architecture}
\end{figure*}

\subsection{Object-Relation Prediction as Set-prediction and Interaction}
A joint \textit{object-relation} graph generation requires searching from a pairwise combinatorial space of the maximum number of expected nodes. Hence, a naive joint-learning for \textit{object-relations} requires $\mathcal{O}(N^2)$ number of tokens for $N$ number of objects. This is computationally intractable because self-attention is quadratically-complex to the number of tokens. We overcome this combinatorially challenging formulation with a carefully engineered inductive bias. The inductive bias, in this case, is to exploit learned pair-wise interactions among $N$ [\texttt{obj}]-tokens and combine refined pair-wise [\texttt{obj}]-tokens with an additional ${(N+1)}^{\text{th}}$ token, which we refer to as  [\texttt{rln}]-token. One can think of the [\texttt{rln}]-token as a query to pair-wise object interaction.

The [\texttt{rln}]-token captures the additional context of pair-wise interactions among all valid predicted classes. In this process, related objects are incentivized to have a strong correlation in an embedding space of, and unrelated objects are penalized to be dissimilar. The [\texttt{rln}]-token attends to all $N$ [\texttt{obj}]-tokens along with contextualized image features that enrich its local pairwise and global image reasoning. Finally, we classify a pair-wise relation by combining the pair-wise [\texttt{obj}]-tokens with the [\texttt{rln}]-token. Thus, instead of $\mathcal{O}(N^2)$ number of tokens, we only need $N+1$ tokens in total. These consist of $N$ [\texttt{obj}]-tokens and one [\texttt{rln}]-token. This novel formulation allows relation detection with a marginally increased cost compared to one-stage object detection.

Here, one could present a two-fold argument: 1) There is no need for an extra token as one could directly classify joint pairwise [\texttt{obj}]-tokens. 2) Instead of one single [\texttt{rln}]-token, one could use as many as the number of possible object-pairs. To answer the first question, we argue that relations can be viewed as a higher order topological entity compared to objects. Thus, to capture inter-dependencies among the relations the model requires additional expressive capacity, which can be shared among the objects. The [\texttt{rln}]-token reduces the burden on the [\texttt{obj}]-tokens by specializing exclusively on the task of relation prediction. Moreover, [\texttt{obj}]-tokens can also attend to the [\texttt{rln}]-token and exploit a global semantic reasoning. This hypothesis has been confirmed in our ablation. For the second question, we argue that individual tokens for all possible object-pairs will lead to a drastic increase in the decoder complexity, which may results in computationally intractability.

\subsection{\methodname}
\vspace{-0.2em}
The \textit{\methodname~}architecture is intuitive and without any bells and whistles, see Fig. \ref{fig:architecture}. We have four main components: a backbone, a transformer, an object detection head and a relation prediction head. In the following, we describe each of the components and the set-based loss formulations specific to joint \textit{object-relation} graph generation in detail.
\vspace{-0.5em}
\paragraph{\textbf{Backbone:}}
Given the input image $I$, a convolutional backbone \cite{he2016deep} extracts features $\boldsymbol{f}_I\in \mathbb{R}^{D_f\times \text{\# emb}}$, where $D_f$ is the spatial dimensions of the features and \text{\# emb} denotes embedding dimension. Further, this feature dimension is reduced to $d_{\text{emb}}$, the embedding dimension of the transformer, and flattened by its spatial size. The new sequential features coupled with the sinusoidal positional encoding \cite{bello2019attention} produce the desired sequence which is processed by the encoder.
\vspace{-0.5em}
\paragraph{\textbf{Transformer:}}
We use a transformer encoder-decoder architecture with deformable attention \cite{zhu2020deformable}, which considerably speeds up the training convergence of DETR by exploiting spatial sparsity of the image features.
\vspace{-0.5em}
\paragraph{Encoder:} Our encoder remains unchanged from \cite{zhu2020deformable}, and uses multi-scale deformable self-attention. We use a different number of layers based on each task's requirement, which is specified in detail in the supplementary material.
\vspace{-0.5em}
\paragraph{Decoder:} We use $N+1$ tokens for the joint \textit{object-relation} task as inputs to the decoder, where $N$ represents the number of [\texttt{obj}]-tokens preceded by a single [\texttt{rln}]-token. Contextualized image features from the encoder serve as the second input of our decoder. In order to have a tractable computation and to leverage spatial sparsity, we use deformable cross-attention between the joint tokens and the image features from the encoder. The self-attention in the decoder remains unchanged.
The [\texttt{obj}]-tokens and [\texttt{rln}]-token go through a series of multi-hop information exchanges with other tokens and image features, which gradually builds a hierarchical object and relational semantics. Here, [\texttt{obj}]-tokens learn to attend to specific spatial positions, whereas the [\texttt{rln}]-token learns how objects interact in the context of their semantic or global reasoning.
\vspace{-0.5em}
\paragraph{\textbf{Object Detection Head:}}
The object detection head has two components. The first one is a stack of fully connected network or  multi layer-perceptron (MLP), which regresses the location of objects, and the second one is a single layer classification module. For each refined [\texttt{obj}]-token $o^i$, the object detection head predicts an object class $\tilde{v}^i_{\text{cls}}=\boldsymbol{W}_{\text{cls}}(o^i)$ and an object location $\tilde{\boldsymbol{v}}^i_{\text{box}}=\text{MLP}_{\text{box}}(o^i), \tilde{\boldsymbol{v}}^i_{\text{box}}\in [0,1]^{2\times d}$ in parallel, where $d$ represents the image dimension, $\boldsymbol{W}_{\text{cls}}$ is the classification layer, and $\text{MLP}_{\text{box}}$ is an MLP. We use the normalized bounding box co-ordinate for scale invariant prediction. Note that for the spatio-structural graph, we create virtual objects around each node's center by assuming an uniform bounding box with a normalized width of $\Delta x$.
\vspace{-0.5em}
\paragraph{\textbf{Relation Prediction Head:}}
In parallel to the object detection head, the input of the relation head, given by a pair-wise [\texttt{obj}]-token and a shared [\texttt{rln}]-token, is processed as
     $\tilde{e}^{ij}_{\text{rln}} =\text{MLP}_{\text{rln}}(\{o^i, r, o^j\}_{i\neq j})$.
Here, $r$ represents the refined [\texttt{rln}]-token and $\text{MLP}_{\text{rln}}$ a three-layer fully-connected network headed by layer normalization \cite{ba2016layer}. In the case of directional relation prediction (e.g., scene graph), the \textit{ordering} of the object token pairs $\{o^i, r, o^j\}_{i\neq j}$ determines the direction $i\rightarrow j$. Otherwise (e.g., road network, vessel graph), the network is trained to learn object token \textit{order} invariance as well.

\subsection{Loss Function}
\label{sec:loss_function}

For object detection, we utilize a combination of loss functions. We use two standard box prediction losses, namely the $\ell_1$ regression loss $(\mathcal{L}_{\text{reg}}$) and the generalized intersection over union loss ($\mathcal{L}_{\text{gIoU}}$) between the predicted $\tilde{\boldsymbol{v}}_{\text{box}}$ and ground truth $\boldsymbol{v}_{\text{box}}$ box coordinates. Besides, we use the cross-entropy classification loss ($\mathcal{L}_{\text{cls}}$) between the predicted class $\tilde{v}_{\text{cls}}$ and the ground truth class $v_{\text{cls}}$.

\paragraph{Stochastic Relation Loss:}
In parallel to object detection, their pair-wise relations are classified with a cross-entropy loss. Particularly, we only use predicted objects that are assigned to ground truth objects by the Hungarian matcher. When two objects have a relation, we refer to their relation as a `valid'-relation. Otherwise, the relation is categorized as `background'. Since `valid'-relations are highly sparse in the set of all possible permutations of objects, computing the loss for every possible pair is burdensome and will be dominated by the `background' class, which may hurt performance. To alleviate this, we randomly sample three `background'-relations for every `valid'-relation. From sampled `valid'- and `background'-relations, we obtain a subset $\mathcal{R}$ of size $M$, where $\mathcal{R} \subseteq \Myperm[N]{2}$. To this end, $\mathcal{L}_{\text{rln}}$ represents a classification loss for the predicted relations in $\mathcal{R}$. The total loss for simultaneous \textit{object-relation} graph generation is defined as:
\begin{eqnarray}
 \mathcal{L}_{\text{total}}= &\sum_{i=1}^{N}\left[ \mathds{1}_{v_{\text{cls}}^i \notin \varnothing} (  \lambda_{\text{reg}}\mathcal{L}_{\text{reg}}(\boldsymbol{v}_{\text{box}}^i,\tilde{\boldsymbol{v}}_{\text{box}}^i)  + \lambda_{\text{gIoU}}\mathcal{L}_{\text{gIoU}}(\boldsymbol{v}_{\text{box}}^i,\tilde{\boldsymbol{v}}_{\text{box}}^i) )\right]\nonumber \\
 +& \lambda_{\text{cls}}\sum_{i=1}^{N}\mathcal{L}_{\text{cls}}(v_{\text{cls}}^i,\tilde{v}_{\text{cls}}^i) + \lambda_{\text{rln}}\sum_{\{i,j\}\in \mathcal{R}}\mathcal{L}_{\text{rln}}(e^{ij}_{\text{rln}}, \tilde{e}^{ij}_{\text{rln}})
\end{eqnarray}
where $\lambda_{\text{reg}}, \lambda_{\text{gIoU}}, \lambda_{\text{cls}}$ and $\lambda_{\text{rln}}$ are the loss functions specific weights.

\section{Experiments}
\label{sec:exp}

\subsection{Datasets}
We conducted experiments on four public datasets for the tasks of road network generation (20 US cities \cite{he2020sat2graph}, Toulouse \cite{belli2019image}), 3D synthetic vessel graph generation \cite{tetteh2020deepvesselnet}, and scene-graph generation (Visual Genome \cite{krishna2016visual}). The road and vessel graph generation datasets are spatio-structural with a binary node and edge classification task, while the scene-graph generation dataset is spatio-sematic and has 151 node classes and 51 edge classes, including `background' class.

\begin{table}[t!]
\centering
\scriptsize
\caption{Brief summary of the datasets used in our experiment. For more details regarding dataset preparation, please refer to supplementary material.}
\label{tab:dataset}
\begin{tabular}{l|c|c|c|c|c|ccc}
\hline
\multirow{2}{*}{Dataset}  & \multicolumn{4}{c|}{Description} & \multicolumn{3}{c}{Data Split} \\
 & Edge Type & 2D/3D   & Image Type & Image Size & Train   & Val & Test   \\
 \midrule
Toulouse \cite{belli2019image}                  & Undirected    & 2D    & Binary      & 64x64    & 80k   & 12k  & 19k  \\
20 US cities\cite{he2020sat2graph}              & Undirected    & 2D    & RGB         & 128x128  & 124k  & 13k  & 25k  \\
Synthetic vessel \cite{tetteh2020deepvesselnet} & Undirected    & 3D    & Grayscale   & 64x64x64 & 54k   & 6k   & 20k  \\ 
Visual Genome \cite{krishna2016visual}          & Directed      & 2D    & RGB         & 800x800  & 57k   & 5k   & 26k  \\ 
\bottomrule 
\end{tabular}

\end{table}

\subsection{Evaluation Metrics}
Given the diversity of tasks at hand, we resort to widely-used task-specific metrics. Following is a brief description, while details can be found in the supplementary material. 
For \textit{Spatio-Structural Graphs}, we use four different metrics to capture spatial similarity alongside the topological similarity of the predicted graphs. 1) \textit{Street Mover Distance (SMD)}\cite{belli2019image} computes a Wasserstein distance between predicted and ground truth edges; 2) \textit{TOPO Score}\cite{he2020sat2graph} includes precision, recall, and F-1 score for topological mismatch; 3) \textit{Node Detection} yields mean average precision (mAP) and mean average recall (mAR) for the node; and  4)\textit{Edge Detection} yields mAP and mAR for the edges.
For \textit{Spatio-semantic Graphs}, the scene graph detection (SGDet) metric is the most challenging and appropriate for joint object-relation detection tasks, because it does not need apriori knowledge about object location or class label. Hence, we compute recall$@K$, mean-recall$@K$, and no-graph constraint (ng)-recall$@K$ for $K=\{20,50,100\}$ on the SGDet task following Zellers et al. \cite{zellers2018neural}. Further, we evaluate the quality of object detection using average precision, AP@50 (IoU=0.5) \cite{lin2014microsoft}.

\subsection{Results}
\paragraph{Spatio-structural Graph Generation: }
\begin{table*}[ht!]
\centering
\scriptsize
\caption{Quantitative comparison of \methodname~with the different baselines for undirected graph generation datasets. \methodname~achieves a near-perfect solution for the Toulouse dataset and improves the results on the 20 US Cities dataset over baseline models. \methodname~translates a similar trend in 3D and significantly outperforms the heuristic-based approach on the synthetic vessel dataset.}
\label{table:results1}
\begin{tabular}{ll|r|rrr|rr|rr}
    \toprule
    \multirow{2}{*}{Dataset} & \multirow{2}{*}{Model} & \multicolumn{4}{c|}{Graph-level Metrics} &  \multicolumn{2}{c|}{Node Det.} &  \multicolumn{2}{c}{Edge Det.}\\
    \cmidrule{3-10}
     &  & SMD $\downarrow$ & Prec. $\uparrow$  & Rec. $\uparrow$ & F1 $\uparrow$ &  mAP $\uparrow$ & mAR $\uparrow$ & mAP $\uparrow$ & mAR $\uparrow$\\
    \midrule
    \multirow{5}{*}{\shortstack[l]{Toulouse\\(2D)}} 
    & RNN \cite{belli2019image}         & 0.04857  & 65.41 & 57.52 & 61.21 & 0.50 & 5.01  & 0.21 & 2.56  \\
    & GraphRNN \cite{belli2019image}    & 0.02450  & 71.69 & 73.21 & 72.44 & 1.34 & 4.15 & 0.34 & 1.01   \\
    & GGT \cite{belli2019image}         & 0.01649  & 86.95 & 79.88 & 83.26 & 2.94 & 13.31  & 1.62 & 9.75 \\
    \cmidrule{2-10}
    & \methodname~                      & \textbf{0.00012}   & \textbf{99.76} & \textbf{98.99} & \textbf{99.37} &  \textbf{94.59} & \textbf{96.76} & \textbf{83.30} & \textbf{89.87} \\
    \midrule
    \multirow{5}{*}{\shortstack[l]{20 US Cities\\(2D)}} 
    & RoadTracer\cite{bastani2018roadtracer}    & N.A.  & 78.00 & 57.44 & 66.16 & N.A. & N.A. & N.A. & N.A. \\
    & Seg-DRM\cite{mattyus2017deeproadmapper}   & N.A.  & 76.54 & 71.25 & 73.80 & N.A. & N.A. & N.A. & N.A. \\
    & Seg-Orientation\cite{batra2019improved}   & N.A.  & 75.83 & 68.90 & 72.20 & N.A. & N.A. & N.A. & N.A. \\
    & Sat2Graph \cite{he2020sat2graph}          & N.A.  & 80.70 & 72.28 & 76.26 & N.A. & N.A. & N.A. & N.A. \\
    \cmidrule{2-10}
    & \methodname~ & 0.04939 &  \textbf{85.28} & \textbf{77.75} & \textbf{81.34}  & 29.25 & 42.84 & 33.19 & 13.45 \\
    \midrule
    \midrule
    \multirow{2}{*}{\shortstack[l]{Synthetic Vessel\\(3D)}} & U-net\cite{ronneberger2015u}+heuristics & 0.01982  & N/A & N/A & N/A & 18.94 & 29.81 & 17.88 & 27.63  \\
    \cmidrule{2-10}
    & \methodname~ & \textbf{0.01107} & N/A & N/A & N/A & \textbf{78.51} & \textbf{84.34} & \textbf{78.10} & \textbf{82.15} \\
    \bottomrule
    \end{tabular}\\
    \scriptsize{*N.A. indicates scores are not readily available. $\dagger$ N/A indicates that the metric is not applicable.}

\end{table*}

In spatio-structural graph generation, both correct graph topology and spatial location are equally important. Note that the objects here are represented as points in 2D/3D space. For practical reasons, we assume a hypothetical box of $\Delta x=0.2$ around these points and treat these boxes as objects.

The Toulouse dataset poses the least difficulty as we can predict a graph from a binary segmentation image. We notice that existing methods perform poorly. Our method improves the SMD score by three orders of magnitude. All other metrics, such as TOPO-Score (prec., rec., and F-1), indicate near-optimal topological accuracy of our method. At the same time, our performance in node and edge mAP and mAR is vastly superior to all competing methods. 
For the more complex 20 U.S. cities dataset, we observe a similar trend. Note that due to the lack of existing scores from competing methods (SMD, mAP, and mAR), we only compare the TOPO scores, which we outperform by a significant margin. However, when compared to the results on the Toulouse dataset, \methodname~yields lower node detection scores on the 20 U.S. cities dataset, which can be attributed to the increased dataset complexity. Furthermore, the edge detection score also deteriorates. This is due to the increased proximity of edges, i.e., parallel roads.

For 3D data, such as vessels, no learning-based comparisons exist. Hence, we compare to the current best practice \cite{shit2021cldice}, which relies on segmentation, skeletonization, and heuristic pruning of the dense skeleta extracted from the binary segmentation \cite{drees2021scalable}. The purpose of pruning is to eliminate redundant neighboring nodes, which is error-prone due to the voxelization of the connectivity, leading to poor performances. Table \ref{table:results1} clearly depicts how our method outperforms the current method. Importantly, we find that our method effortlessly translates from 2D to 3D without major modifications. Moreover, our 3D model is trained end-to-end from scratch without a pre-trained backbone. To summarize, we propose the first reliable learning-based 3D spatio-structural graph generation method and show how it outperforms existing 2D approaches by a considerable margin. 
\paragraph{Scene Graph Generation:} We extensively compare our method to numerous existing methods, which can be grouped based on three concepts. One-stage methods, two-stage methods utilizing only image features, and two-stage methods utilizing extra features. Importantly, \methodname~represents a one-stage method without the need for extra features. 
We find that \methodname~outperforms all one stage methods in Recall and ng-Recall despite using a simpler backbone. In terms of mean-Recall, a metric addressing dataset bias, we outperform \cite{liu2021fully} and our contemporary \cite{cong2022reltr} @50 and perform close to \cite{cong2022reltr} @20.

\begin{table*}[t!]
\scriptsize
\centering
\label{tab:vg_result}
\caption{Quantitative results of \methodname~in comparison with state-of-the-art methods on the Visual Genome dataset. \methodname~achieves new one-stage state-of-the-art results and bridges the performance gap with two-stage models, while reducing model complexity and inference time significantly without the need for any extra features (e.g., glove vector, knowledge graph, etc.). Importantly, \methodname~outperforms two-stage models that previously reported mean-Recall@100 and ng-Recalls. Note that `-' indicates that the corresponding results are not available to us.}
\begin{adjustbox}{width=1.0\textwidth,center}
\begin{tabular}{c|c|c|rrr|rrr|rrr|r|r|r}
 \toprule
 \multicolumn{2}{c|}{\multirow{2}*{Method}}&Extra & \multicolumn{3}{c|}{ Recall}&\multicolumn{3}{c|}{ mean-Recall}&\multicolumn{3}{c|}{ ng-Recall}& AP & \#param & \multirow{2}*{FPS $\uparrow$} \\
  \multicolumn{2}{c|}{} & Feat. & @20 &@50 & @100 &@20 &@50 & @100 &@20 &@50 & @100 & @50 &(M)$\downarrow$ &  \\
  \hline 
 \multirow{9}*{\rotatebox[origin=c]{90}{Two-Stage}}
& MOTIFS \cite{zellers2018neural}    &\cmark & 21.4 & 27.2 & 30.5 & 4.2 & 5.7 & 6.6 & - &3 0.5 & 35.8 & 20.0 & 240.7 & 6.6\\
& KERN \cite{chen2019knowledge}      &\cmark & 22.3 & 27.1 & - & - & 6.4 & - & - & \textcolor{blue}{\textbf{30.9}} & 35.8 & 20.0 & 405.2 & 4.6 \\
& GPS-Net \cite{lin2020gps}          &\cmark & 22.3 & 28.9 & \textcolor{blue}{\textbf{33.2}} & 6.9 & 8.7 & \textcolor{blue}{\textbf{9.8}} & - & - & - & - & - & - \\
& BGT-Net  \cite{dhingra2021bgt}     &\cmark & 23.1  & 28.6 & 32.2 & - & - & 9.6 & - & - & - & - & - & - \\
& RTN \cite{koner2020relation}       &\cmark & 22.5  & 29.0 & 33.1 & - & - & - & - & - & - & - & - & - \\
& BGNN \cite{li2021bipartite}        &\cmark & \textcolor{blue}{\textbf{23.3}} &  \textcolor{blue}{\textbf{31.0}} & - & \textcolor{blue}{\textbf{7.5}} & \textcolor{blue}{\textbf{10.7}} & - & - & - & -  &  \textcolor{blue}{\textbf{29.0}} & 341.9 & 2.3\\
& GB-Net \cite{zareian2020bridging}  &\cmark & -  & 26.3 & 29.9 & - & 7.1 & 8.5 & - & 29.3 & 35.0 & - & -  & - \\
& \cellcolor{blue!10}{IMP{+}} \cite{xu2017scene} & \cellcolor{blue!10}{\xmark} & 14.6 & 20.7 & 24.5 & 2.9 & 3.8 & 4.8 & - & 22.0 & 27.4 & 20.0 &  \textcolor{blue}{\textbf{203.8}} &  \textcolor{blue}{\textbf{10.0}}\\
& \cellcolor{blue!10}{G-RCNN}\cite{yang2018graph} & \cellcolor{blue!10}{\xmark} & - & 11.4 & 13.7 & - & - &-  &- & 28.5 &  \textcolor{blue}{\textbf{35.9}} & 23.0 & - & -\\
\hline
\hline
 \multirow{4}*{\rotatebox[origin=c]{90}{\shortstack[l]{One-\\Stage}}}
& \cellcolor{blue!10}{FCSGG} \cite{liu2021fully}&\cellcolor{blue!10}{\xmark} & 16.1 & 21.3 &- & 2.7 & 3.6 &-  & 16.7 & 23.5 & 29.2 & \textbf{28.5} & 87.1  & ${8.3}^*$\\
& \cellcolor{blue!10}{RelTR}  \cite{cong2022reltr} &\cellcolor{blue!10}{\xmark} & 20.2 & 25.2 & - & \textbf{5.8} & 8.5 & -  & - & - & - & 26.4 & \textbf{63.7} & 16.6\\
\cmidrule{2-15}
& \cellcolor{blue!10}{\methodname~} &\cellcolor{blue!10}{\xmark} & \textbf{22.2}  & \textbf{28.4} & \textbf{31.3} & 4.6 & \textbf{9.3} & \textbf{10.7}  &\textbf{22.9} &\textbf{31.2} &\textbf{36.8} & 26.3 & 92.9 & $\textbf{18.2}^*$\\
\bottomrule
\end{tabular}
\end{adjustbox}
\scriptsize{\#param are taken from \cite{cong2022reltr}. * Frame-per-second (FPS) is computed in Nvidia GTX 1080 GPU.}

\end{table*}

In terms of object detection performance, we achieve an AP@50 of $26.3$, which is close to the best performing one- and two-stage methods, even though we use a simpler backbone. Note that the object detection performance varies substantially across multiple backbones and object detectors. For example, BGNN \cite{li2021bipartite} uses X-101FPN, FCSGG \cite{liu2021fully} uses HRNetW48-5S-FPN, whereas \methodname~and its contemporary RelTR \cite{cong2022reltr} use a simple ResNet50 \cite{he2016deep} backbone. 

Comparing our \methodname~to two-stage models, we outperform all models that use no extra features in all metrics. Moreover, we perform almost equal to the remaining two-stage models, which use powerful backbones \cite{li2021bipartite}, bi-label graph resampling \cite{li2021bipartite}, custom loss functions \cite{lin2020gps}, and  extra features such as word \cite{koner2020relation} or knowledge graph embeddings \cite{chen2019knowledge}. Therefore, we can claim that we achieve competitive performances without custom loss functions or extra features while using significantly fewer parameters. We also achieve much faster processing times, measured in frames per second (FPS)(see Table \ref{tab:vg_result}). For example, BGNN \cite{li2021bipartite}, which was the top performer in a number of metrics, requires three times more parameters and is an order of magnitude slower than our method.

\begin{figure*}[ht!]
    \centering
    \scriptsize
    \includegraphics[width=0.95\textwidth]{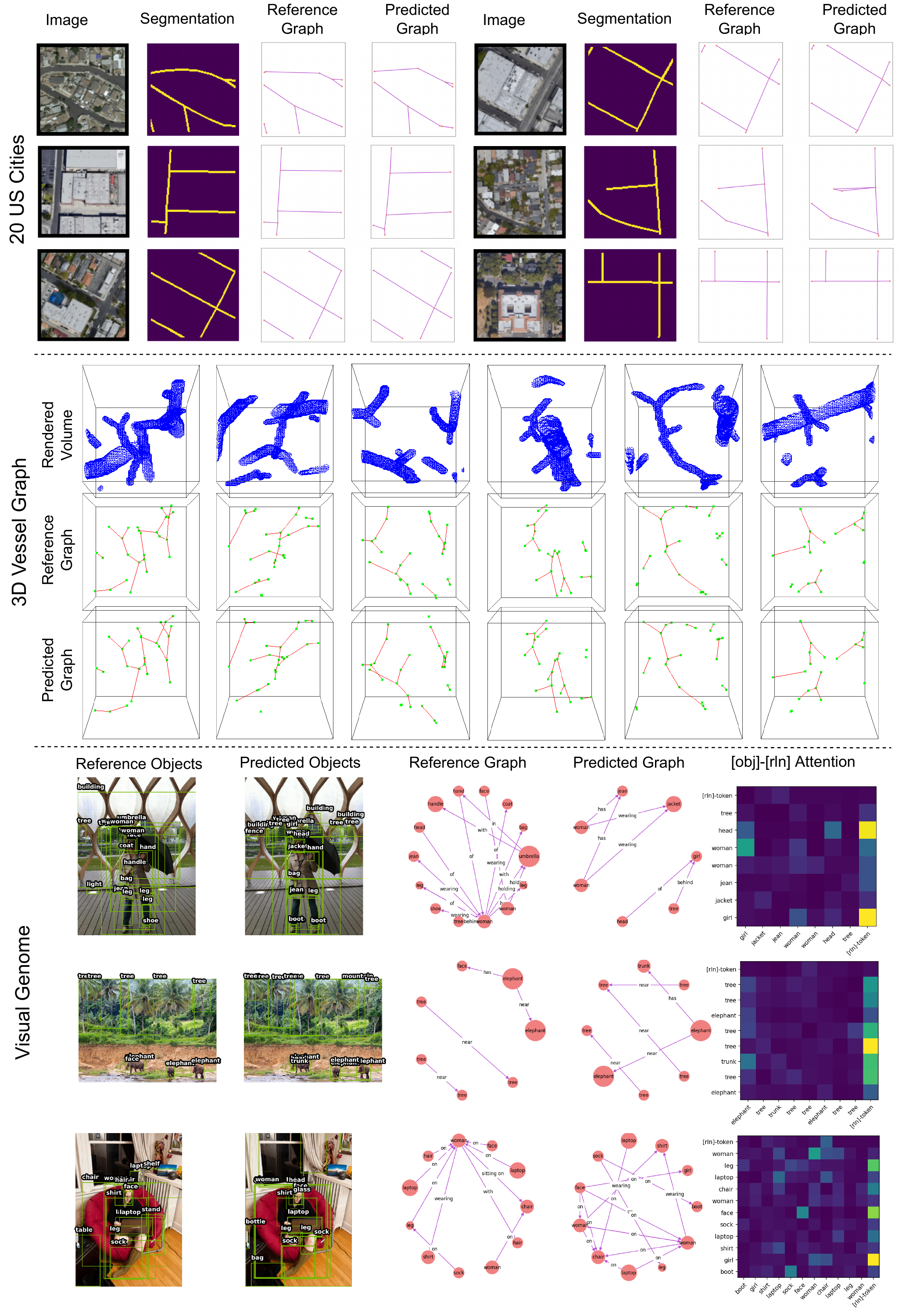}

    \caption{Qualitative results (better viewed zoomed in) from road-network, vessel-graph, and scene-graph generation experiments. Across all datasets, we observe that \methodname~is able to produce correct results. The segmentation map is given for better interpretability of road network satellite images. For vessel-graphs, we surface-render the segmentation of corresponding greyscale voxel data. For scene graphs, we visualize the attention map between detected [\texttt{obj}]-tokens and [\texttt{rln}]-token, which shows that the [\texttt{rln}]-token actively attends to objects that contribute to relation formation.}
    \label{fig:qualitative}

\end{figure*}

Fig. \ref{fig:qualitative} shows qualitative examples for all datasets used in our experiments.
Qualitative and quantitative results from both spatio-structural and spatio-semantic graph generation demonstrate the efficiency of our approach and the importance of simultaneously leveraging [\texttt{obj}]-tokens and the [\texttt{rln}]-token. \methodname~achieves benchmark performances across a diverse set of image-to-graph generation tasks suggesting its wide applicability and scalability.

\subsection{Ablation Studies}
In our ablation study, we focus on two aspects. First, how the [\texttt{rln}]-token and relation-head guide the graph generation; second, the effect of the sample size in training transformers from scratch. We select the complex 3D synthetic vessel and Visual Genome datasets for the ablation. Further ablation experiments can be found in the supplementary material.

In Table \ref{table:vg_ablation}, we evaluate the importance of the [\texttt{rln}]-token and different relation-head types. First, we train def-DETR only for object detection as proposed in \cite{carion2020end,zhu2020deformable}, second, we evaluate \methodname~w/ and w/o  [\texttt{rln}]-token and use a linear relation classification layer (models w/o the [\texttt{rln}]-token use only concatenated pair-wise [\texttt{obj}]-tokens for relation classification). Third, we replace the linear relation head with an MLP and repeat the same w/ and w/o [\texttt{rln}]-tokens.

\begin{table*}[t!]
\scriptsize
\centering
\hspace*{-.1em} 
\begin{minipage}[b]{0.46\textwidth}
\caption{\scriptsize Ablation on the [\texttt{rln}]-token and relation head type on Visual Genome. We observe [\texttt{rln}]-token significantly improves relation prediction for both types of relation heads. Importantly, the improvement is larger for the linear classifier than for the MLP.
}
\label{table:vg_ablation}
\begin{adjustbox}{width=1.0\textwidth,center}
\begin{tabular}{lclc|ccc}
\toprule
\multirow{2}{*}{Model} 
& \multirow{2}{*}{\shortstack[l]{[\texttt{rln}]-\\token}} & \multirow{2}{*}{\shortstack[l]{Rel.\\Head}} & \multirow{2}{*}{\shortstack[l]{AP\\@50}} & \multicolumn{3}{c}{SGDet Recall} \\ 
\cline{4-7} 
 &  &   &   & @20          & @50          & @100         \\ 
 \midrule
 def-DETR    & N/A      &  N/A    &  26.4    & N/A       & N/A    & N/A      \\
 \midrule
\methodname  & \xmark   & Linear  & 24.1     & 16.6      & 22.0   & 25.2     \\ 
\methodname  & \cmark   & Linear  & 25.3     & 20.1      & 25.4   & 28.3     \\ 
\midrule
\methodname  & \xmark   & MLP     &  26.0    & 19.2      & 26.4   & 29.4     \\
\methodname  & \cmark   & MLP     & 26.3     & 22.2      & 28.4   & 31.3     \\ 
\bottomrule
\end{tabular}
\end{adjustbox}
\end{minipage}
\hspace*{0.5em}
\begin{minipage}[b]{0.52\textwidth}
\scriptsize
\caption{\scriptsize Ablation on the [\texttt{rln}]-token and train-data size on synthetic vessel. We observe that [\texttt{rln}]-token significantly improves both node and edge detection. Additionally, the scores improves with increased train-data size, suggesting further room for improvement by training on more data.}
\label{table:vessel_ablation}
\begin{adjustbox}{width=1.0\textwidth,center}
\begin{tabular}{lcl|c|cc|cc}
    \toprule
     \multirow{2}{*}{Model}
     & \multirow{2}{*}{\shortstack[l]{[\texttt{rln}]-\\token}}
     &\multirow{2}{*}{\shortstack[l]{Train\\Data}} & \multirow{2}{*}{SMD} &  \multicolumn{2}{c|}{Node Det.} &  \multicolumn{2}{c}{Edge Det.}\\
    \cline{5-8}
     &  &  & &  mAP & mAR & mAP & mAR\\
    \midrule
    def-DETR    & N/A    & 100\% & N/A    & 77.5 & 83.5 & N/A  & N/A  \\
    \midrule
    \methodname & \xmark & 100\% & 0.0129 & 75.5 & 81.6 & 76.3 & 80.4 \\
    \midrule
    \methodname & \cmark & 25\%  & 0.0138 & 17.0 & 32.1 & 11.5 & 28.3 \\
    \methodname & \cmark & 50\%  & 0.0124 & 39.2 & 53.5 & 33.3 & 48.9 \\
    \methodname & \cmark & 100\% & \textbf{0.0110}  &  \textbf{78.5} & \textbf{84.3} & \textbf{78.1} & \textbf{82.1} \\
    \bottomrule
    \end{tabular}
\end{adjustbox}

\end{minipage}

\end{table*}

We observe that a linear relation classifier w/o [\texttt{rln}]-token is insufficient to model the mutual relationships among objects and diminishes the object detection performance as well. In contrast, we see that the [\texttt{rln}]-token significantly improves performance despite using a linear relation classifier. Using an MLP instead of a linear classifier is a better strategy whereas the \methodname~w/ [\texttt{rln}]-token shows a clear benefit. Unlike the linear layer, we hypothesize that the MLP provides a separate and adequate embedding space to model the complex semantic relationships for [\texttt{obj}]-tokens and our [\texttt{rln}]-token.

From ablation on 3D vessel (Table \ref{table:vessel_ablation}), we draw the same conclusion that [\texttt{rln}]-token significantly improve over \methodname~w/o [\texttt{rln}]-token. Further, a high correlation between performance and train-data size indicates additional room for improvement by increasing the sample size while training from scratch.

\subsection{Limitations and Outlook:} In this work, we only use bipartite object matching, and future work will investigate graph-based matching \cite{rolinek2020deep}. Additionally, incorporating recent transformer-based backbones, i.e., Swin-transformer \cite{liu2021swin} could further boost the performance without compromising the simplicity.

\section{Conclusion}
Extraction of structural- and semantic-relational graphs from images is the key for image understanding. We propose \methodname,~a unified single-stage model for direct \textit{image-to-graph} generation. Our method is intuitive and easy to interpret because it is devoid of any hand-designed components. We show consistent performance improvement across multiple \textit{image-to-graph} tasks using \methodname compared to previous methods; all while being substantially faster and using fewer parameters which reduce energy consumption. \methodname~opens up new possibilities for efficient integration of a  \textit{image-to-graph} models to downstream applications in an end-to-end fashion. We believe that our method has the potential to shed light on many previously unexplored domains and can lead to new discoveries, especially in 3D. 

\section*{Acknowledgement}
Suprosanna Shit is supported by TRABIT (EU Grant: 765148). Bjoern Menze gratefully acknowledges the support of the Helmut Horten Foundation.

\bibliographystyle{plain}
\bibliography{main}
\newpage
\appendix
\section{Transformer and Deformable-DETR}
The core of a transformer \cite{vaswani2017attention} is the attention mechanism. Let us consider an image feature map $\boldsymbol{f}_I$, the $q^{th}$ query with associated features $\boldsymbol{f}_q$ and $k^{th}$ key with associated image features $\boldsymbol{f}^k_I$. One can define the multi-head attention for $M$ number of heads and $K$ number of key elements as
\begin{align}
    \operatorname{MultiHeadAttn}\left(\boldsymbol{f}_{q}, \boldsymbol{f}_I\right)=\sum_{m=1}^{M} \boldsymbol{W}_{m}\left[\sum_{k =1}^{K} \boldsymbol{A}_{m q k} \cdot \boldsymbol{W}_{m}^{\prime} \boldsymbol{f}^k_{I}\right]\nonumber
\end{align}
where $\boldsymbol{W}_{m}^{\prime}$ and $\boldsymbol{W}_{m}$ are learnable weights. The attention weights $\boldsymbol{A}_{m q k} \propto \exp \left\{\frac{\boldsymbol{f}_{q}^{\top} \boldsymbol{W}_{m}^{\prime\prime \top} \boldsymbol{W}^{\prime\prime\prime}_{m} \boldsymbol{f}^k_{I}}{\sqrt{d_k}}\right\}$ are normalized as $\sum_{k =1}^K \boldsymbol{A}_{m q k}=1$, where $\boldsymbol{W}^{\prime\prime}_{m}, \boldsymbol{W}^{\prime\prime\prime}_{m}$ are also learnable weights and $d_k$ is the temperature parameter. To differentiate position of each element uniquely, $\boldsymbol{f}_{q}$ and $\boldsymbol{f}_{I}$ are given a distinct positional embedding.

In our work, we use the multi scale deformable attention \cite{zhu2020deformable} for $L$ number of level as 
\begin{equation}
    \text{MSDefAttn}(\boldsymbol{f}_q,\boldsymbol{x}_q,\{\boldsymbol{f}_I^l\}_{l=1}^L) = \sum_{m=1}^{M}\boldsymbol{W}_m\left[\sum_{l=1}^{L}\sum_{k=1}^{K}\boldsymbol{A}_{mlqk}\cdot\boldsymbol{W}_{m}^{'}\boldsymbol{f}_I^l(\phi_l(\boldsymbol{x}_q) + \Delta \boldsymbol{x}_{mlqk})\right]\nonumber
\end{equation}
where $\phi_l$ rescales the normalized reference point coordinates appropriately in the corresponding image space.

\section{Dataset} Here we describe the individual datasets used in our experimentation in detail. We also elaborate on generating train-test sets for our experiments. For 20 U.S. Cities and 3D synthetic vessel we extract overlapping patches from large images. This provides us a large enough sample size to train our \methodname~from scratch. Since, a DETR like architecture is not translation invariant because of learned [\texttt{obj}]-tokens in the decoder, extracting overlapping patches drastically increases the effective sample size within a limited number of available images.

\subsection{Toulouse Road Network}
The Toulouse Road Network dataset \cite{belli2019image} is based on publicly available satellite images from Open Streetmap and consists of semantic segmentation images with their corresponding graph representations. For our experiments we use the same split as in the original dataset paper with 80,357 samples in the training set, 11,679 samples in the validation set, and 18,998 samples in the test set \cite{belli2019image}.
\subsection{20 U.S. Cities Dataset}
\begin{figure}[t]
    \centering
    \includegraphics[width=1.0\textwidth]{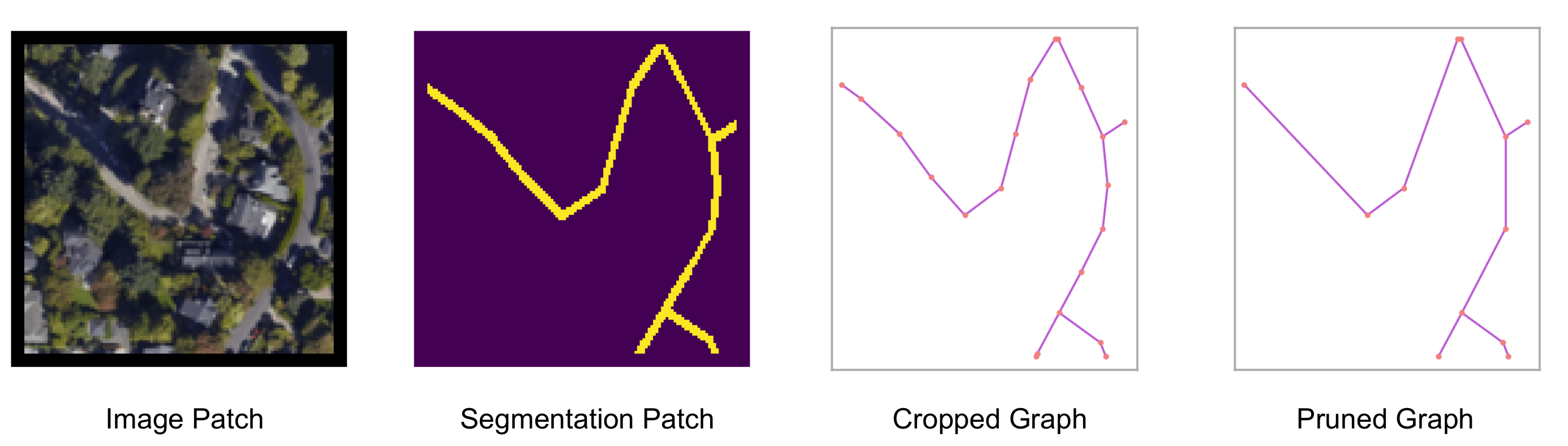}
    \vspace{-2em}
    \caption{Preprocessing steps for the 20 U.S. Cities dataset. The same steps are followed in the 3D Synthetic Vessel dataset curation.}
    \label{fig:prerprocess}
    \vspace{-1.5em}
\end{figure}

For the 20 U.S. Cities dataset \cite{he2020sat2graph}, there are 180 images with a resolution of 2048x2048. We select 144 for training, 9 for validation, and 27 for testing. From those images, we extract overlapping patches of size 128x128 to construct the final train-validation-test split. We crop the RGB image and the corresponding graph followed by a node simplification. Following Belli et al. \cite{belli2019image}, we prune the dense nodes by computing the angle between two road-segments at each node of degree 2 and only keep a node if the road curvature is less than 160 degrees. This allows eliminating redundant nodes and simplifying the graph prediction task. Fig. \ref{fig:prerprocess} illustrates the pruning process.

\subsection{3D Synthetic Vessels}
Our synthetic vessel dataset is based on publicly available synthetic images generated in Tetteh et al. \cite{tetteh2020deepvesselnet}. In this dataset, the ground truth graph was generated by \cite{schneider2012tissue} and from that, corresponding voxel-level semantic segmentation data was generated. Grey valued data was obtained by adding different noise levels to the segmentation map. Specifically, we train on greyscale "images" and their corresponding vessel graph representations, where each node represents a bifurcation point, and the edges represent their connecting vessels. The whole dataset contains 136 3D volumes of size 325x304x600. First, we choose 40 volumes to create a train and validation set and next pick 10 volumes for the test set. From this, we extract overlapping patches of size 64x64x64 to construct the final train-validation-test set. Similar to the 20 U.S. cities dataset, we prune nodes having degree 2 based on the angle between two edges. 
\subsection{Visual Genome}
Visual Genome is one of the largest scene graph datasets consisting of 108,077 natural images \cite{krishna2016visual}. However, the original dataset suffers from multiple annotation errors and improper bounding boxes. Lu et al. \cite{lu2016visual} proposed a refined version of Visual Genome with the most frequent occurring 150 objects classes and 50 relation categories. It also proposed its own train/val/test splits and is the most widely used data-split \cite{zellers2018neural,koner2020relation,lin2020gps,liu2021fully} for SGG. For fair comparison, we only train on the Visual Genome dataset and do \textbf{not} use any pre-training.

\section{Metrics Details}

\begin{figure}[t]
    \centering
    \includegraphics[width=1.0\textwidth]{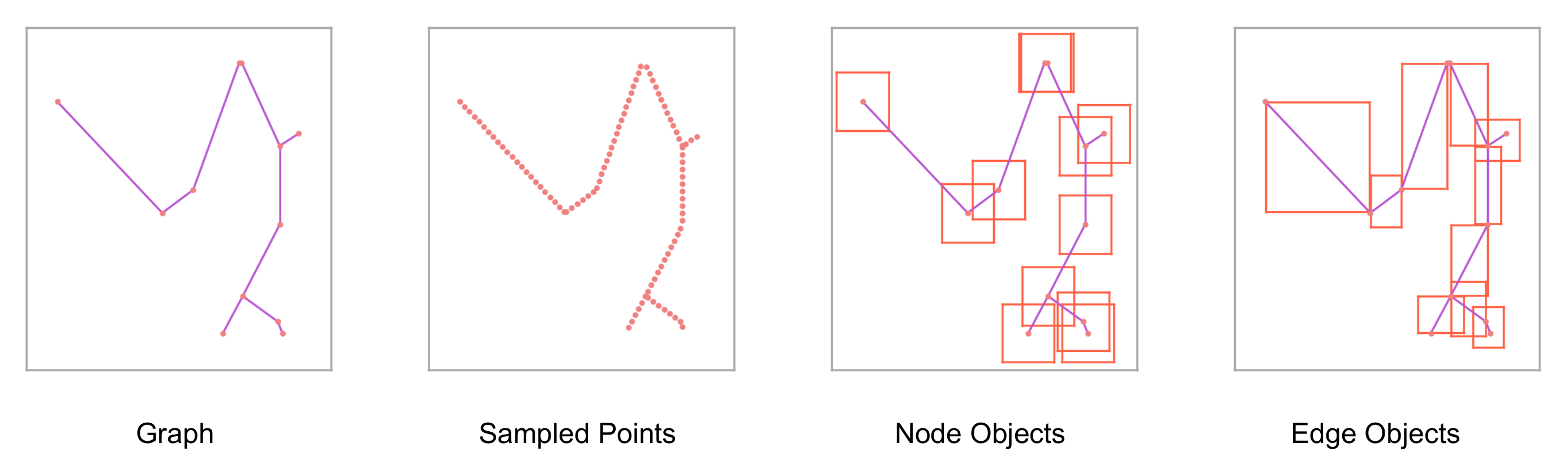}
    \vspace{-2em}
    \caption{Sampled points, node objects and edge objects for computing different spatio-structural graph metrics. The same notion is used for 3D graphs.}
    \label{fig:metrics}
\end{figure}

\paragraph{Metrics for Spatio-Structural Graph:} We use three different kinds of metrics to capture spatial similarity alongside the topological similarity of the predicted graphs. The graph-level metrics include; 1) \textit{Street Mover Distance (SMD):} SMD\cite{belli2019image} compute Wasserstein distance between the uniformly sampled fixed number of points (See Fig. \ref{fig:metrics}) from the predicted and ground truth edges; and 2) \textit{TOPO Score:} TOPO Score\cite{he2020sat2graph} computes precision, recall, and F-1 score for topological mismatch in terms of the false-positive and false-negative topological loop. Alongside, we use 3) \textit{Node Detection:} For this, we report mean average precision (mAP) and mean average recall (mAR) over a threshold range [0.5,0.95,0.05] for node box prediction. Similarly, we use 4) \textit{Edge Detection:} We compute the mAP and mAR for the edge in the same way as above. The edge boxes are constructed from the center points of two connecting nodes (See Fig. \ref{fig:metrics}). For vertical and horizontal edges we assume an hypothetical width of 0.15 to avoid objects with near zero width.
\paragraph{Metrics for Spatio-Semantic Graph:} 
We evaluate \methodname~ on the most challenging Scene Graph Detection(SGDet) metrics and its variants. Unlike other scene graph metrics like Predicate Classification (PredCls) or Scene Graph classification (SGCls) , SGDet does not use apriori information on class label or object spatial position and does not rely on complex RoI-align based spatial features. SGDet jointly measures the predicted boxes (with $50\%$ overlaps) class labels of an object, and relation labels. The variants of SGDet  include 1) \textit{Recall:} Recall at the different K (\@20, \@50 and \@100) of predicted relation that reflects overall relation prediction performance, 2) \textit{Mean-Recall:} mean-Recall computes mean of each relation class-wise recall that reflects the performance under the relational imbalance or long-tailed distribution of relation  class, 3) \textit{ng-Recall:} ng-Recall is recall w/o graph constraints on the prediction, which takes the top-k predictions instead of just the top-1. Additionally, we use 4) AP@50: Average precision at 50\% threshold of IOU reflects an average object detection performance. 

\section{Model Details}

\begin{table}[ht!]
\scriptsize
\centering
\caption{The model parameters used in \methodname~experiments across the various datasets. Specifically, we list details on the backbone and the transformer's number of layers, feature dimension and other details.}
\label{tab:full_model}

\begin{tabular}{l|l|r|r|r|r|r} 
\toprule
\multirow{2}{*}{DataSet} & \multirow{2}{*}{Backbone} & \multicolumn{4}{c|}{Transformer} & \multirow{2}{*}{MLP Dim} \\ 
\cline{3-6}
                        &                 & Enc. Layer  & Dec. Layer & \# [\texttt{obj}]-tokens &$d_{\text{emb}}$  & \\ 
\midrule
Toulouse                &  ResNet-50      & 4           & 4          & 20              &         256        & 512\\
20 US cities            &  ResNet-101     & 4           & 4          & 80              & 512      & 1024 \\ 
Synth Vessel            &  SE-Net      & 4           & 4          & 80              &       256          & 1024\\
Visual Genome           &  ResNet-50      & 6           &  6         & 200             & 512             & 2048\\
\bottomrule
\end{tabular}
\end{table}
Table \ref{tab:full_model}, describes the backbone and important parameters of the \methodname. We experiment with different ResNet backbones to show the flexibility of our \methodname. In order to reduce energy consumption, we use the lighter ResNet50 for most 2D datasets. For the 3D experiment, we used Squeeze-and-Excite Net \cite{hu2018squeeze}.
We used the number of encoder and decoder layers and the number of [\texttt{obj}]-tokens in the increasing order of dataset complexity. We find that four transformer layers and 20 [\texttt{obj}]-tokens suffice for Toulouse, while we need four transformer layers and 80 [\texttt{obj}]-tokens are required for 20 U.S. cities and synthetic vessel datasets. We need 6 layers of transformer and 200 [\texttt{obj}]-tokens for the visual genome. The ablation on the number of transformer layers and number of [\texttt{obj}]-tokens are shown in the next section.

\section{Training Details}

\begin{table}[ht!]
\centering
\scriptsize
\caption{A list of the important set of parameters used  in \methodname~for respective training. Furthermore, we list the weights for bipartite matching costs and training losses.}
\label{tab:full_training}
\begin{tabular}{l|r|r|r|r|r|r|r|r|r|r} 
\toprule
\multirow{2}{*}{DataSet} & \multirow{2}{*}{\shortstack[r]{Batch\\Size}} & \multirow{2}{*}{\shortstack[r]{Learning\\rate}} & \multirow{2}{*}{Epoch} & \multicolumn{3}{c|}{Cost Coeff.} & \multicolumn{4}{c}{Loss Coeff.}  \\ 
\cline{5-11}
&         &       &    &    cls & reg & gIoU & $\lambda_{\text{reg}}$ & $\lambda_{\text{gIoU}}$ & $\lambda_{\text{cls}}$ & $\lambda_{\text{rln}}$  \\ 
\midrule
Toulouse        &    64    &   $10^{-4}$   &    50   &   2    &   5   & 0 & 5    & 2    & 2    & 1     \\
20 US cities    &    32    & $10^{-4}$  & 100   &    3   &    5   &  0  & 5     & 2    & 3    & 4     \\
3D Vessel Net   & 48     &   $10^{-4}$    &  100     & 2 &     5      &   0   &  2   & 3    &  3   & 4     \\
Visual Genome   & 16     & $10^{-4}$  & 25    &   3    &  2     & 3    &  2  & 2    & 4    & 6    \\
\bottomrule
\end{tabular}
\arrayrulecolor{black}
\end{table}
Table. \ref{tab:full_training}, summarizes some principal parameters we use in the training. We use AdamW optimizer with a step learning rate. For scene graph generation, we use the prior statistical distribution or frequency-bias \cite{zellers2018neural} of relation for each subject-object pair. To minimize the data imbalance for a relation label present in the Visual Genome, we use log-softmax distribution \cite{lin2020gps} to soften the frequency bias. Finally, we add this distribution with the predicted relation distribution from the relation head. For the spatio-structural dataset, we set the cost coefficient for the GIoU in the bipartite matcher to be zero because we assume 0.2 widths of the normalized box for each node. Hence, $\ell_1$ cost is sufficient to consider for the spatial distances.

\section{More Ablation Studies on [\texttt{obj}]-tokens and Transformer}
\begin{figure*}[ht!]
    \centering
    \scriptsize
    \includegraphics[width=1.0\textwidth]{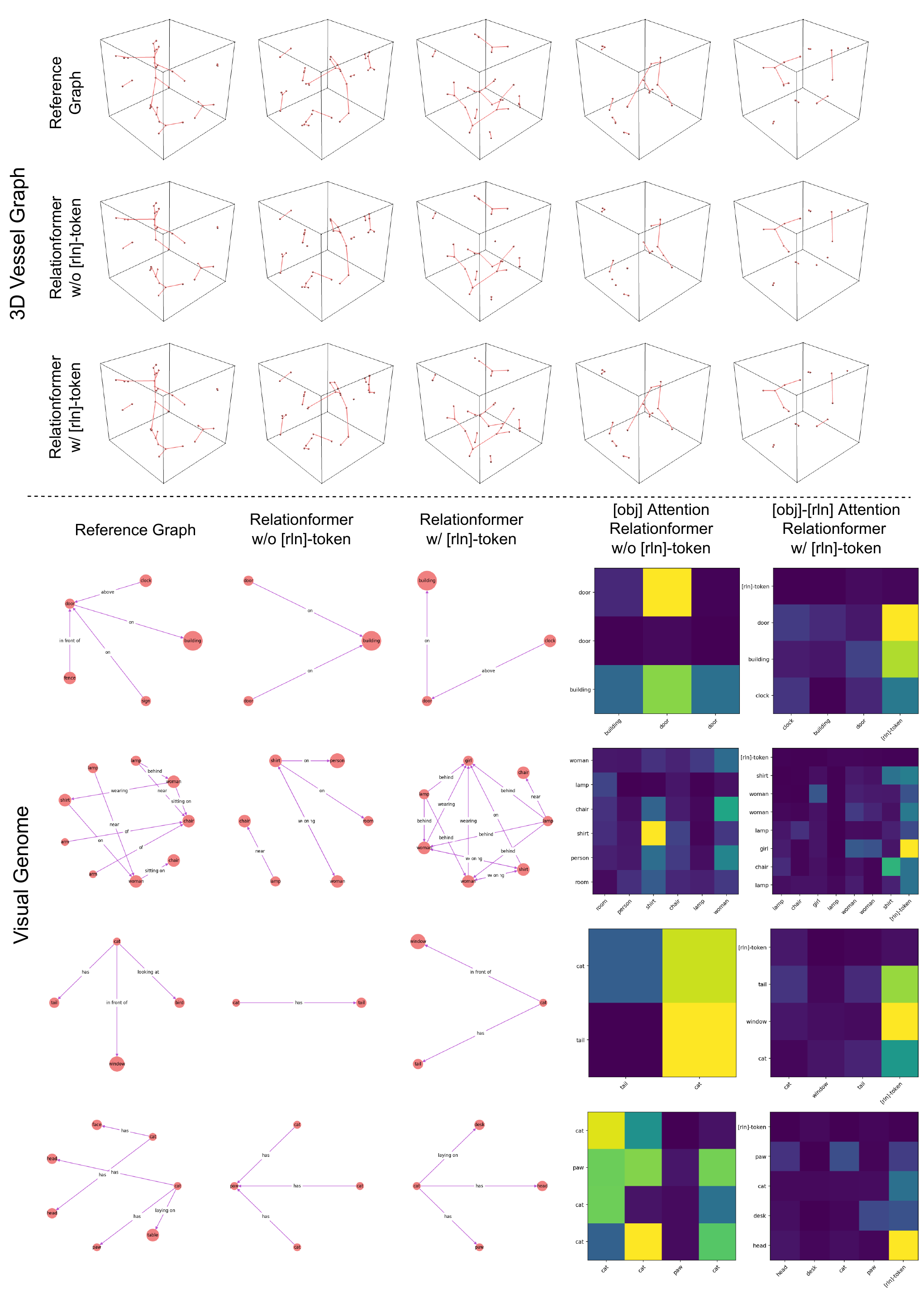}
    \vspace{-1em}
    \caption{Typical qualitative results (please zoom in) from our ablation on the synthetic vessel-graph and visual genome datasets. We observe that \methodname~w/o [\texttt{rln}]-token is missing vessel edges while \methodname~w/ [\texttt{rln}]-token produces correct edges. For visual genome, we can see w/o [\texttt{rln}]-token the [\texttt{obj}]-tokens have to carry extra burden for relation prediction and sometimes fail to incorporate the global relation. However, the inclusion of [\texttt{rln}]-token provides an additional path to flow relation information that benefits the joint object and relation detection.}
    \vspace{-1em}
    \label{fig:ablation}
    \vspace{-1em}
\end{figure*}

We conduct two more ablation studies on Visual Genome for analyzing the influence of [\texttt{obj}]-tokens and optimal number of layers in transformer for the joint graph generation. Furthermore Figure. \ref{fig:ablation} gives additional insight how [\texttt{rln}]-token is beneficial for joint object-relation graph.

\begin{table}[ht!]
\scriptsize
\centering
\begin{minipage}[t]{0.46\textwidth}
\caption{Impact of the [\texttt{obj}]-tokens on joint object and relation detection.}
\label{table:vg_objtkn}
\begin{adjustbox}{width=1.0\textwidth,center}
\begin{tabular}{lc|ccc}
\toprule
\multirow{1}{*}{\#[\texttt{obj}]-tokens}            & AP@50           & R@20   & R@50   & R@100   \\ 
\midrule 
75             & 25.1             & 20.6          & 26.1      & 29.5   \\
100                  & 25.8                  & 21.1          & 27.4              & 30.6         \\
\textbf{200(ours)}             & 26.3             & 22.2         & 28.4         & 31.3        \\
300            & 26.3             & 21.9         & 27.9         & 31.0  \\
\bottomrule      
\end{tabular}
\end{adjustbox}
\end{minipage}
\begin{minipage}[t]{0.46\textwidth}
\caption{Impact of the transformer's layers on joint object and relation detection}
\label{table:vg_transformer layer}
\begin{adjustbox}{width=1.0\textwidth,center}
\begin{tabular}{lc|ccc}
\toprule
\multirow{1}{*}{\# layer}            & AP@50           & R@20   & R@50   & R@100   \\ 
\midrule 
4             & 24.6             & 20.5          & 26.5       & 28.8  \\
5                  & 25.2                  & 21.0          & 27.2              & 29.9         \\
\textbf{6(ours)}             & 26.3             & 22.2         & 28.4         & 31.3        \\

\bottomrule      
\end{tabular}
\end{adjustbox}
\end{minipage}
\end{table}
\par As shown in Table \ref{table:vg_objtkn}, it can be observed that increasing [\texttt{obj}]-tokens does increase object and relation detection performance. However, it becomes relatively stable with increasing object quarries. DETR-like architectures rely on an optimal number of [\texttt{obj}]-tokens to balance positive and negative simple which also helps in object detection as observed in \cite{carion2020end}. Thus, in a joint object and relation prediction, a gain might come from optimal number [\texttt{obj}]-tokens, as relation prediction is linearly co-related to object detection performance. It demonstrates that joint object and relation detection can perfectly coexist without hurting the object detection performance. Instead, it can exploit [\texttt{obj}]-tokens enriched with global relational reasoning for efficient relation extraction. 

During the ablation with transformer layers, we observe decreasing number of transformer layers shows an initial gain in object and relation detection. However, they lead to early plateau and inferior performance as depicted in table \ref{table:vg_transformer layer}. One intuitive reason is that with less parameter and insufficient contextualization \methodname~quickly learn the initial biases present in both object and relation detection and failed to learn the complex global scenario. We use the same number of layers for both encoder and decoder.

\section{Qualitative Results}
\begin{figure*}[ht!]
    \centering
    \scriptsize
    \includegraphics[width=1.0\textwidth]{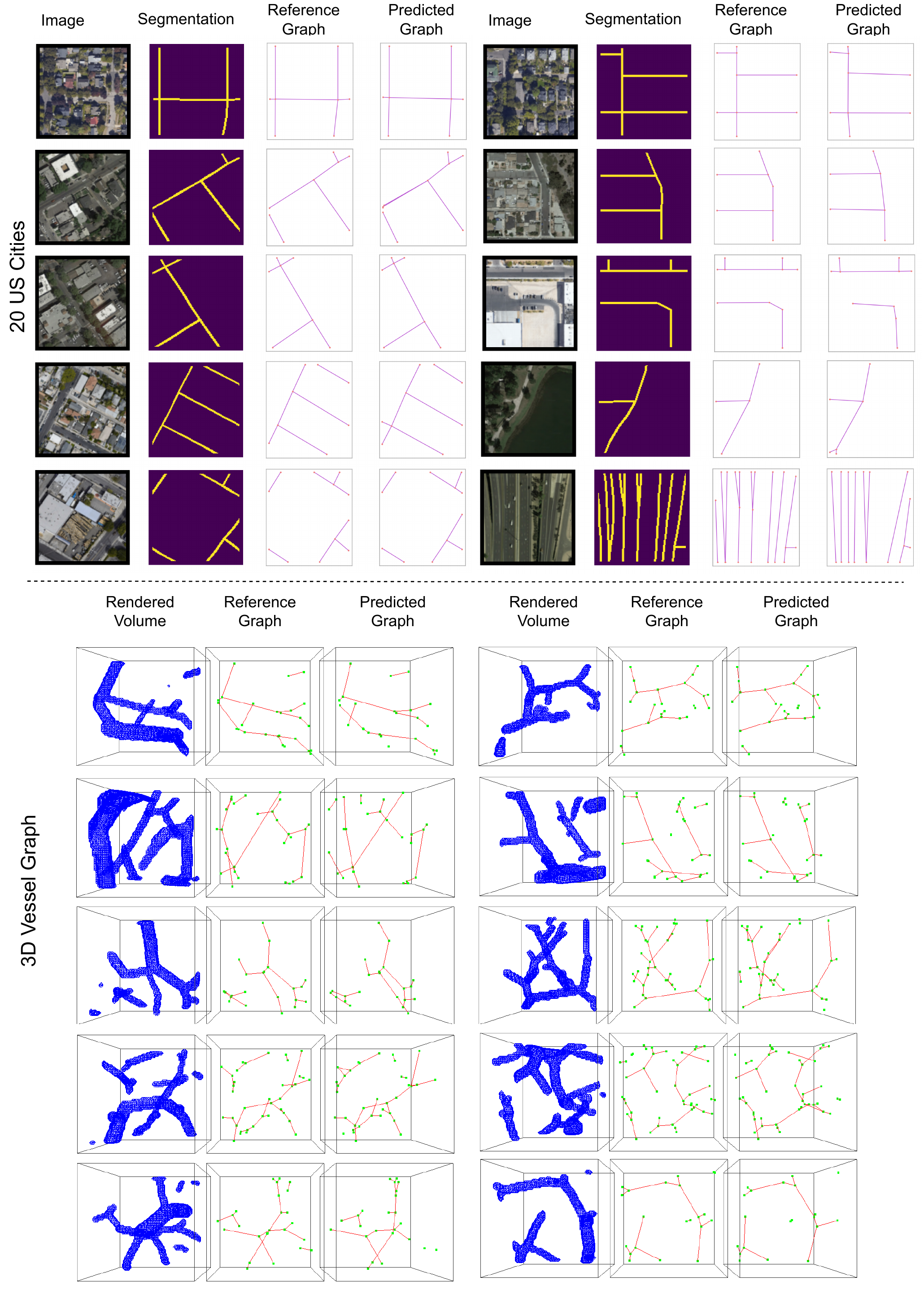}
    
    \caption{Qualitative results (please zoom in) for the 20 US cities road-network and synthetic vessel-graph experiments. We observe that \methodname~is able to produce correct results. The segmentation map is given for better interpretability of road network satellite images. For vessel-graphs, we surface-render the segmentation of the corresponding greyscale voxel data.}
    \label{fig:qualitative1}
    \vspace{-2em}
\end{figure*}

\begin{figure*}[ht!]
    \centering
    \scriptsize
    \includegraphics[width=1.0\textwidth]{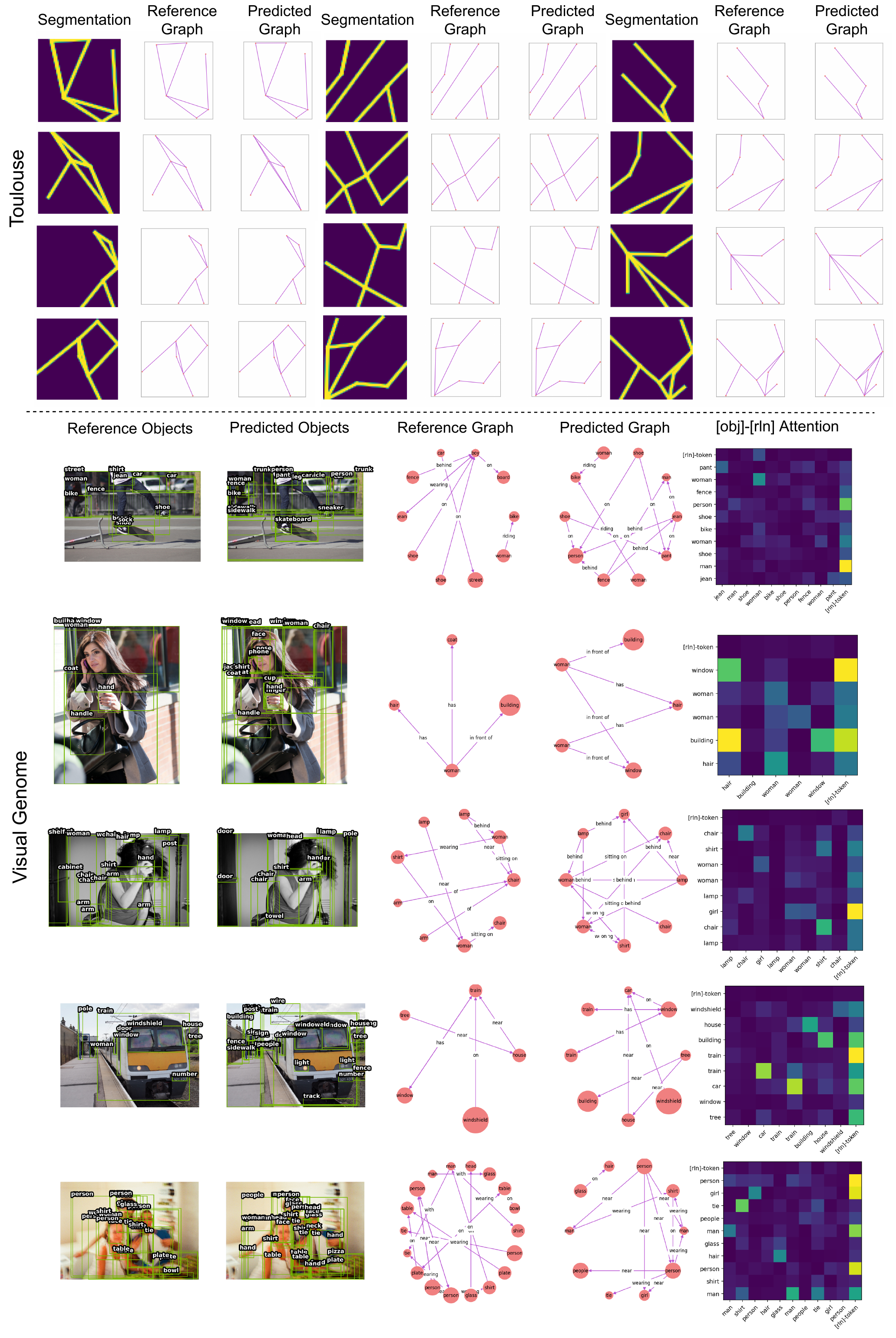}
    \vspace{-2em}
    \caption{Qualitative results (please zoom in) from the Toulouse road-network and scene-graph generation experiments. For both datasets, we observe that \methodname~is able to generate an accurate graph. For scene graphs, we visualize the attention map between detected [\texttt{obj}]-tokens and [\texttt{rln}]-token, which shows that the [\texttt{rln}]-token actively attends to objects that contribute to relation formation.}
    \label{fig:qualitative2}
    \vspace{-2em}
\end{figure*}

Fig. \ref{fig:qualitative1} and \ref{fig:qualitative2} shows additional qualitative example from our experiments.

\end{document}